# Tracking Direction of Human Movement – An Efficient Implementation using Skeleton

Merina Kundu
St. Xavier's College, Kolkata
30 Mother Teresa Sarani
Kolkata – 700016, India

Dhriti Sengupta
St. Xavier's College, Kolkata
30 Mother Teresa Sarani
Kolkata – 700016, India

Jayati Ghosh Dastidar
St. Xavier's College, Kolkata
30 Mother Teresa Sarani
Kolkata – 700016, India

## ABSTRACT
Sometimes a simple and fast algorithm is required to detect human presence and movement with a low error rate in a controlled environment for security purposes. Here a light weight algorithm has been presented that generates alert on detection of human presence and its movement towards a certain direction. The algorithm uses fixed angle CCTV camera images taken over time and relies upon skeleton transformation of successive images and calculation of difference in their coordinates.

## 1. INTRODUCTION
Identification and tracking of humans in image sequences is a key issue for a variety of application fields, such as, video-surveillance, human-computer interface, animation, and video indexing. Human shape identification and movement detection are challenging tasks in computer image processing, even more when attempted in real time. Object variation, shape variation of the same object, and even changing backgrounds poses several challenges for the same. Here a skeleton based detection [1] and movement algorithm is used that scans over two consecutive frames of a video sequence to detect the motion.

This work is focused on skeleton and its major points and its relative positions in successive picture frames. A static camera takes continuous pictures from a fixed angle with a static background in a controlled environment. Subtracting this background from successive images generates what can be called as the foreground. It consists of only dynamic objects coming into view.

These dynamic objects are then processed. First, its skeleton which is an important shape descriptor is extracted [2]. From this shape it is inferred whether the skeleton is that of a human or not [1]. Comparing the relative coordinates of skeletons so extracted from frames photographed over time gives the movement of the human and its direction. This system automatically generates alerts on detection of such human movement in the aforesaid environment.

This paper is organized as follows. Section 2 describes some related works in this field. Section 3 describes processing of the images for the detection of human beings. Section 4 describes the movement algorithm. In Section 5 some experimental data have been presented. Finally, Section 6 concludes.

## 2. LITERATURE REVIEW
The detection is often used as an initialization step to tracking [3, 4, 5]. Authors in [2, 6, 7, 8, 9, 10, 11, 12, 13] have thoroughly studied the use of skeleton in modeling the object shape for the purpose of matching different shapes, which have already been extracted from the images. Approaches for such type of movement tracking of a human includes that of Song et al in [14] using comparison of certain reference points on a human 2D kinematic model. The same is done by Viola et al. [15] using five filters to detect motion pattern even with a very low resolution image. The model proposed by Fablet and Black [16] compares relative dense flow patterns of human with that of a model. The method employs heavy computational power. Earlier work by Cutler and Davis [17] assumes more system simplification and focuses on periodicity directly from tracked images which works well with low resolution images. Many approaches in this field rely heavily on intermediate representations making detection and system failure plausible. Other related work includes that of Papageorgiou et al [18]. This system detects pedestrians using a support vector machine trained on an over complete wavelet basis. In [19], a Bayesian framework is developed for object localization based on probabilistic modeling of object shape. The most successful tracking methods exploit statistical models of object appearance such as color or grey-level histograms [20], mixture models of color distributions [21], background subtraction [22], edge-based models of object shape [23, 24], etc.

This algorithm is based on skeletons of different objects and their differences. The advantage of this algorithm is its efficiency as predefined ratio is used for detection of humanoid shape in both rigid and non-rigid models. In this paper, the skeleton configuration has been represented using some feature points. Object parts are then identified from these features. In the detection stage, it has been determined whether the overall posture and shape of the skeleton matches that of a humanoid depending on the measurements calculated from these feature points. Next, if the incoming entity is a human being, the relative positions of the human in two consecutive frames are determined to find the direction of movement. Thus, this algorithm is simpler and computationally efficient.

## 3. PROCESSING OF IMAGES TO DETECT HUMAN PRESENCE

### 3.1 Information Collection and Extraction of Incoming Objects

First, static background information is collected with different illumination, without any dynamic object. A stream of real time images, with aforesaid background, ('*frames*'), are fed and used to detect human movement. A frame rate of 10 fps was chosen as it gave a good trade off between observation and reduction of time complexity.

In the next stage any foreign object that had appeared over the static background is identified. To detect any change between real time frames and the static background image two image matrices are compared with a let off level of 5% as deduced from the correlation coefficient [1]. If the result is positive then the static background and the frame are again compared pixel-wise and a new image called **DIFF** is created. Every non-zero difference was set to white, and every zero value was maintained. Thus DIFF gave a black and white image with only the difference highlighted in white [1].

A case study may make the concept simpler and easier to understand. Consider the following images in Figure 1; the first one being the background, and the second one being a frame shot. The correlation was less than 0.95, so a pixel-wise difference was taken and 'DIFF' was found, shown in Figure 1 (c).

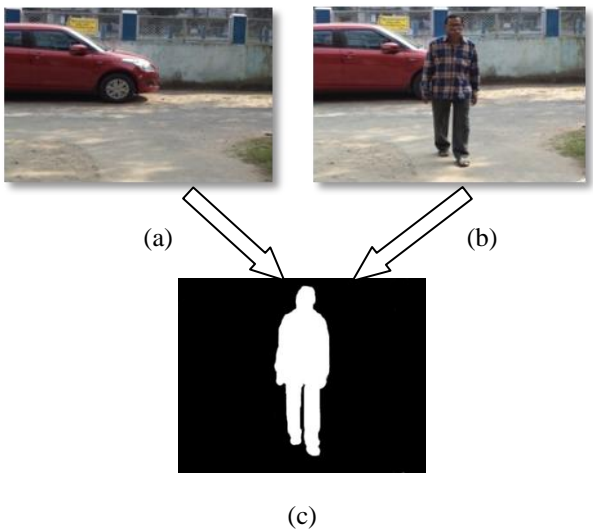

**Figure 1: Two frames, (a) the static background, (b) with a human presence and (c) Dynamic object extraction by Comparison**

### 3.2 Extraction of Feature points

To recognize the object, features of the object are needed. These features preferably should be easy to compute, and easy to manipulate.

A *skeleton*, which is one such feature descriptor, has been used here. It can be described to be a set of points equidistant from the nearest edges of the image [2]. Skeletons of two dimensional objects often show a lot of spurious edges and branches because of image noise. So, the skeleton generation algorithm using 'Discrete Curve Evolution' is used to prune such branches [8].

There are two main advantages of using skeletons for detection of the object class – (1) it emphasizes the geometric and topological properties of the shape; and (2) it retains connectivity in the image (Figure 2).

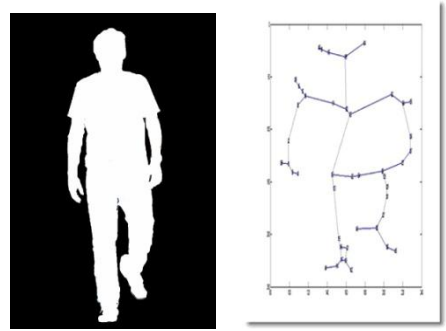

**Figure 2. Example of skeleton of 2D image**

A skeleton point having only one adjacent point has been called an **endpoint**; a skeleton point having more than two adjacent points has been called a **fork point**; and a skeleton segment between two skeleton points has been called a skeleton **branch**. Every point which is not an endpoint or a fork point has been called a **branch point** (Figure 3).

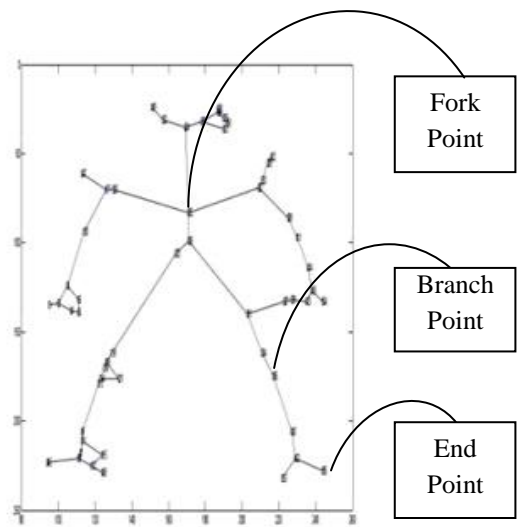

**Figure 3. Skeleton with endpoint, branch point and fork points.**

### 3.3 Human Detection

Once the cleaned up skeleton is found, it is processed to find the shape information. Note that the humanoid shape is primarily determined by its relative positions of limbs (arms and legs), neck and head. Therefore, the points where the skeleton is "broken" into forks are considered to obtain a fair idea of where the neck/head, arms, or legs are. This is very different from other living beings.

Two simple heuristic feature elements of a human skeleton are used. First, its height to width ratio is computed. This gives an estimate of the subject's overall shape. Secondly, the fork points are found. Generally the skeleton of a human should have two major fork points: (1) at the neck; and (2) at the waist. These features are used to generate a **final_score** for the object, indicating whether the object is likely to be a human or not [1].

## 4. MOVEMENT DETECTION

The model-based method of Fablet and Black [16] recovers pose, orientation and position in the image but is computationally heavy. The objective of this work is to find a quick and simple algorithm for a very specific requirement using the skeleton of the detected human. This algorithm is based on the **centre of gravity of the skeleton** deduced by averaging the skeleton's major points such as end points, branch points and fork points. The broad steps of the movement detection algorithm are as follows-

1. Define an array, **cgx_sheet** to store the previous and current cgx values (defined below). Initialize these array cells with 0.

2. For each difference image do the following-

3. Check if the incoming entity is human or not. If human then do the following steps :

4. Add the x-coordinates of all the fork points, branch points and end points and then average it. Call it **cgx.**
   Cgx = Total x-coordinate value of all major points / Total number of all major points

5. Add the y-coordinates of all the fork points, branch points and end points and then average it. Call it **cgy.**
   Cgy = Total y-coordinate value of all major points / Total number of all major points

After finding the **centre of gravity (cgx, cgy)** of the current frame the cgx_sheet array is updated.

6. Update the previous and current value of cgx as follows:
   cgx_sheet[1,1] = cgx_sheet[2,1]
   cgx_sheet[2,1] = cgx

Here cgx_sheet[1,1] and cgx_sheet[2,1] store the previous and current value of cgx which are calculated from the previous and current frames respectively.

If the previous centre of gravity is different from the current centre of gravity then it can be said that the human shape has moved. The movement can be forward or backward, left or right. To find the movement direction the difference of previous and current cgx (cgx_diff) is calculated. The value 5 has been used as a threshold to detect the movement of the human. The value 5 is chosen keeping the frame sizes and the video rate in mind; in this case, a pixel shift of 5 implies a significant movement in real world.

7. If cgx_sheet[1,1] ≠ 0 then do the following

8. Find the difference of the current and previous cgx. Call it **cgx_diff**.
   cgx_diff = cgx_sheet[2,1] – cgx_sheet[1,1]

9. Find the absolute value of cgx_diff.

10. If the absolute value of cgx_diff is greater than 5 then it can be said that the human moves 5 pixels either left or right with respect to the observer else the human moves forward or backward or stands almost in the same position.
    (a) If cgx_diff > +5
        Moving Right, GENERATE ALARM.
    (b) Else cgx_diff < –5
        Moving Left, GENERATE ALARM.

## 5. EXPERIMENT

The software has been developed using MATLAB. It accepts as input live feed from Close Circuit Television Cameras (CCTV) installed at strategic locations. The input from several CCTV cameras deployed in the college premises has been used. The software works by accepting the feed and then splitting it into frames. The frames are then analyzed as described in the preceding sections to detect intrusion, human intrusion and direction of movement if and when a human intrusion occurs.

Here, the results obtained from two data sets are presented: detection and tracking direction of a student's movement. First, the static picture has been shown, followed by the dynamic frames; and finally the difference pictures along with the corresponding skeleton of the dynamic object.

In the first example (Figure 4), first some static pictures have been taken. Then, some real time frames were collected from a video feed with a student in that same place. The human is moving to the right continuously. Here four (4) such frames with dynamic figures are shown. Difference pictures are generated by comparing the static picture to the real time frames. The difference pictures are binary images. The corresponding skeletons of the dynamic objects are then obtained. As the value of cgx_diff is greater than (+5) in each case, it is inferred that the human is moving to the right continuously with respect to the observer. Here cgx_prev is the value of cgx_sheet[1,1] and cgx_new is the value of cgx_sheet[2,1]. Table 1 summarises the data obtained for this example. The graph in Figure 6. (a) shows the positive slope for the change in the value of cgx which is increasing in magnitude over time.

Similarly, in the next example (Figure 5), binary difference pictures are generated by comparing the static picture to the real time frames. The corresponding skeletons of the dynamic objects are obtained. The value of cgx_diff is less than (–5) in each case. So it is inferred that the human is moving to the left continuously with respect to the observer. Table 2 summarises the data obtained for this example. The graph in Figure 6. (b) shows the negative slope for the change in the value of cgx which is decreasing in magnitude over time.

The charts (Figure 6) show the shift of cgx of the skeleton in the consecutive frames with respect to time. An increment of more than +5 in cgx value indicates movement towards right and a decrement of less than –5 in cgx value indicates movement towards left.

## 5.1 Example 1: Moving Right

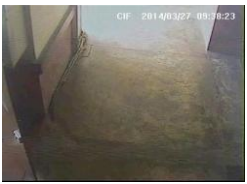

STATIC PICTURE    PLACE: ST XAVIER'S COLLEGE

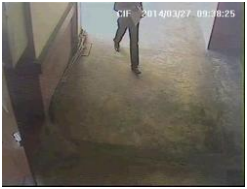 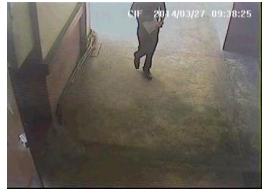 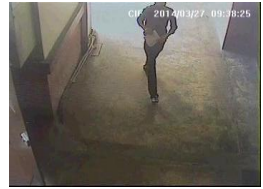 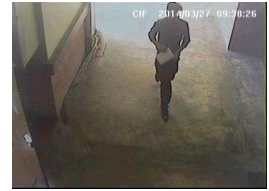

**Dynamic Frame 1**    **Dynamic Frame 2**    **Dynamic Frame 3**    **Dynamic Frame 4**

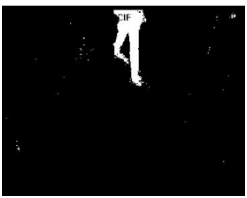 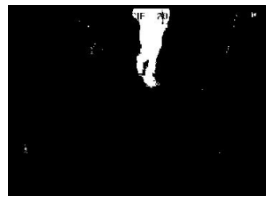 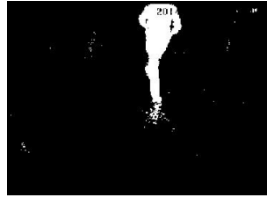 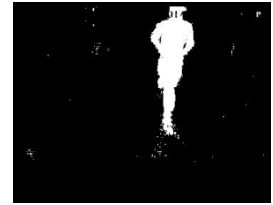

**Difference Picture 1**    **Difference Picture 2**    **Difference Picture 3**    **Difference Picture 4**

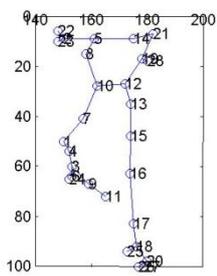 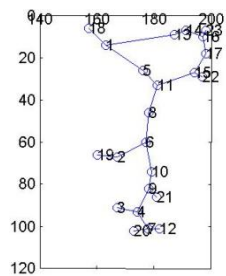 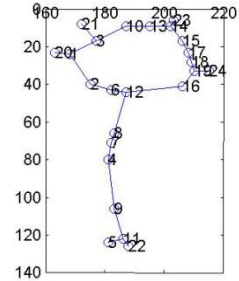 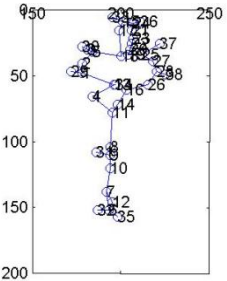

**Skeleton 1**    **Skeleton 2**    **Skeleton 3**    **Skeleton 4**

**Figure 4. Static Image, Dynamic frames, skeletons and scores obtained on 1st set of data**

**Table 1: Results of 1st set of data**

| Frame No. | Final_score | CGX | CGY | Cgx_prev | Cgx_new | Cgx_diff |
|---|---|---|---|---|---|---|
| 1. | 1 | 165.6071 | 47.8929 | 0 | 165.6071 | NIL |
| 2. | 1 | 179.6957 | 50.2174 | 165.6071 | 179.6957 | 14.0885 |
| 3. | 1 | 189.5417 | 45.9167 | 179.6957 | 189.5417 | 9.8460 |
| 4. | 1.4 | 198.1579 | 59.7105 | 189.5417 | 198.1579 | 8.6162 |

## 5.2 Example 2: Moving Left

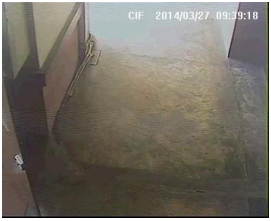

STATIC PICTURE    PLACE: ST XAVIER'S COLLEGE

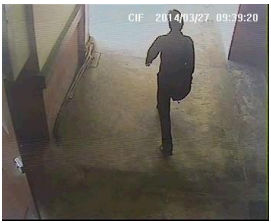 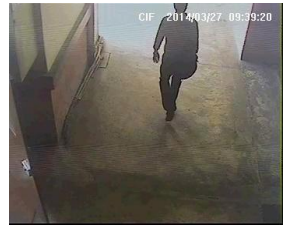 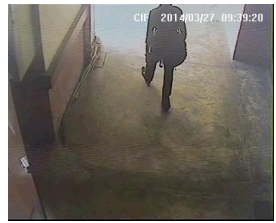 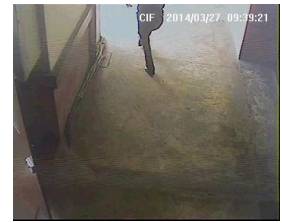

**Dynamic Frame 1**    **Dynamic Frame 2**    **Dynamic Frame 3**    **Dynamic Frame 4**

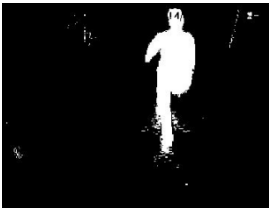 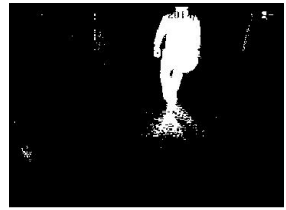 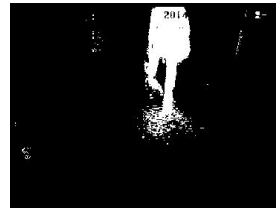 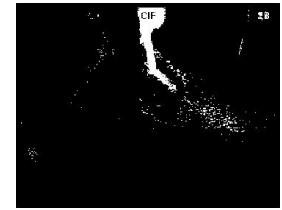

**Difference Picture 1**    **Difference Picture 2**    **Difference Picture 3**    **Difference Picture 4**

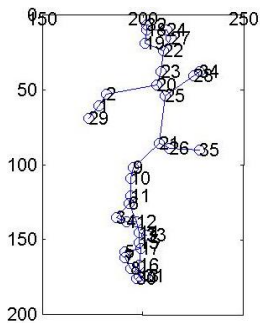 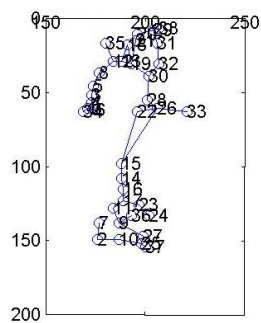 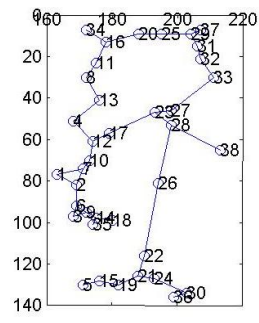 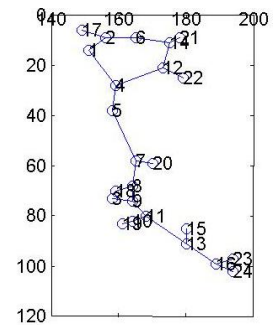

**Skeleton 1**    **Skeleton 2**    **Skeleton 3**    **Skeleton 4**

**Figure 5. Static Image, Dynamic frames, skeletons and scores obtained on 2nd set of data**

**Table 2: Results of 2nd set of data**

| Frame No. | Final_score | CGX | CGY | Cgx_prev | Cgx_new | Cgx_diff |
|---|---|---|---|---|---|---|
| 1. | 1.4 | 200.6286 | 97.1714 | 0 | 200.6286 | NIL |
| 2. | 1.4 | 190.6053 | 75.3421 | 200.6286 | 190.6053 | -10.0233 |
| 3. | 1 | 184.8421 | 68.0789 | 190.6053 | 184.8421 | -5.7632 |
| 4. | 1 | 168.8750 | 53.7917 | 184.8421 | 168.8750 | -15.9671 |

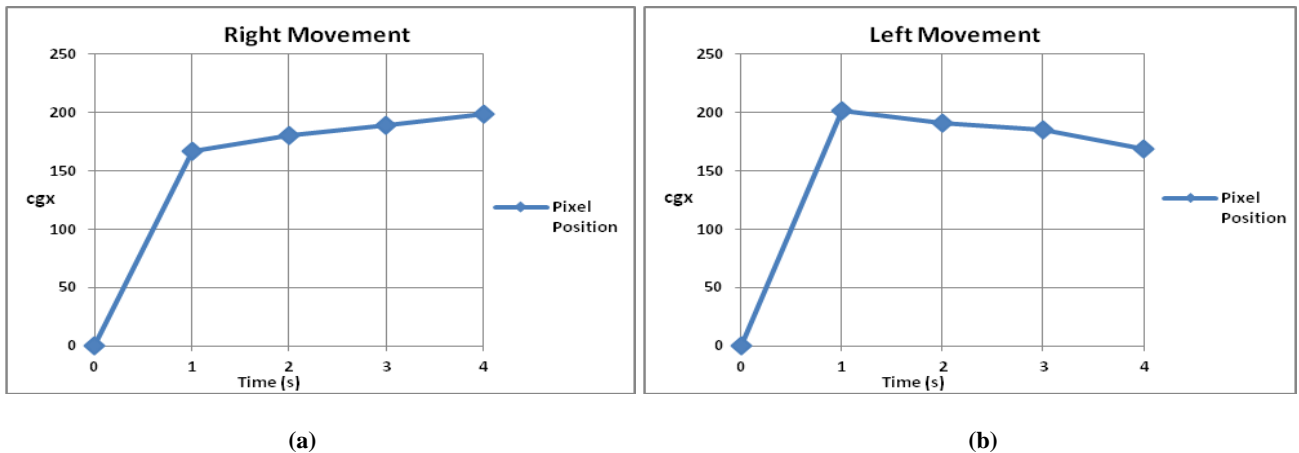

(a)           (b)

**Figure 6. Graphical Representation of Measurements from (a) Experiment 1 and (b) Experiment 2**

## 6. CONCLUSION

The algorithm presented in this paper has been tested exhaustively in different situations involving men and women wearing different clothes; and also on dynamic frames containing non-humans, such as dogs, cats; and rigid objects like cars and boxes. The experiments have been conducted under different illumination levels, at different times of day and at different locations. Promising results were obtained in all cases. This algorithm is capable of tracking human beings in different environments, under different lighting conditions. It is deformity tolerant to a significant extent in the sense that bending or twisting does not affect its ability. Existing generic object tracking algorithms generally have either very limited modeling power, or they are too complicated to learn, and also tends to be computationally expensive. This algorithm is based on some basic arithmetic operations; so it is simple, effective and efficient. The results are encouraging.

There are issues open to further investigation. Very noisy environments (for example, a rapidly changing background) generate erroneous difference images. Sudden sharp change in lighting conditions also introduces artifacts. The ratios which we used were obtained after analyzing average human data; so there are possibilities that some humans will fall out of this range. Also, this algorithm has not been tested on primates or humanoid robots which are close to humans in shape; so its performance is not known in such specific cases. The algorithm judges only lateral movement. Diagonal shift will also introduce a horizontal component and therefore can be detected. Vertical shifts can be caught by looking at cgy. Repetition of the algorithm with successive images may also indicate the path of movement. More experiments on these issues can further enhance the efficiency of the algorithm.

## 7. ACKNOWLEDGMENTS

This work has been supported in part by University Grants Commission (UGC), India as part of a Minor Research Project titled "Automated CCTV Surveillance". The authors would like to extend their gratitude to UGC for funding the work. They would also like to thank the Principal of St. Xavier's College, Kolkata, Rev. Dr. J. Felix Raj, S.J. for all the support he gave for doing research. Finally, they would like to thank the Department of Computer Science, St. Xavier's College, Kolkata, for helping them arrange the infrastructure required (both hardware and software) to conduct the experiments.